\documentclass[conference]{IEEEtran}
\IEEEoverridecommandlockouts
\usepackage{cite}
\usepackage{amsmath,amssymb,amsfonts}
\usepackage{algorithmic}
\usepackage{graphicx}
\usepackage{textcomp}
\usepackage{xcolor}
\usepackage{url}
\usepackage{array,tabularx}
\usepackage{multicol} 
\usepackage{cleveref}
\usepackage{verbatim}
\usepackage{multirow}
\usepackage{booktabs} 

\def\BibTeX{{\rm B\kern-.05em{\sc i\kern-.025em b}\kern-.08em
    T\kern-.1667em\lower.7ex\hbox{E}\kern-.125emX}}
\begin{document}

\title{Evaluation of an Uncertainty-Aware Late Fusion Algorithm for Multi-Source Bird’s Eye View Detections Under Controlled Noise
}

\author{
\IEEEauthorblockN{Maryem Fadili\textsuperscript{1,}\textsuperscript{2}\textsuperscript{*},
Louis Lecrosnier\textsuperscript{2},
Steve Pechberti\textsuperscript{1}, 
Redouane Khemmar\textsuperscript{2}}

\IEEEauthorblockA{\textsuperscript{1}VEDECOM, Versailles, France}

\IEEEauthorblockA{\textsuperscript{2}IRSEEM, Saint-Etienne du Rouvray, France}

\textsuperscript{*}Corresponding author : maryem.fadili@vedecom.fr
}

\maketitle
\begin{abstract}
Reliable multi-source fusion is crucial for robust perception in autonomous systems. However, evaluating fusion performance independently of detection errors remains challenging. This work introduces a systematic evaluation framework that injects controlled noise into ground-truth bounding boxes to isolate the fusion process. We then propose Unified Kalman Fusion (UniKF), a late-fusion algorithm based on Kalman filtering to merge Bird’s Eye View (BEV) detections while handling synchronization issues. Experiments show that UniKF outperforms baseline methods across various noise levels, achieving up to 3× lower object's positioning and orientation errors and 2× lower dimension estimation errors, while maintaining near-perfect precision and recall between $99.5\%$ and $100\%$.
\end{abstract}

\begin{IEEEkeywords}
Autonomous driving, Collaborative perception, Object detection, Sensor fusion
\end{IEEEkeywords}

\begin{figure*}[h]
  \centering  
   \includegraphics[width=0.9\linewidth]{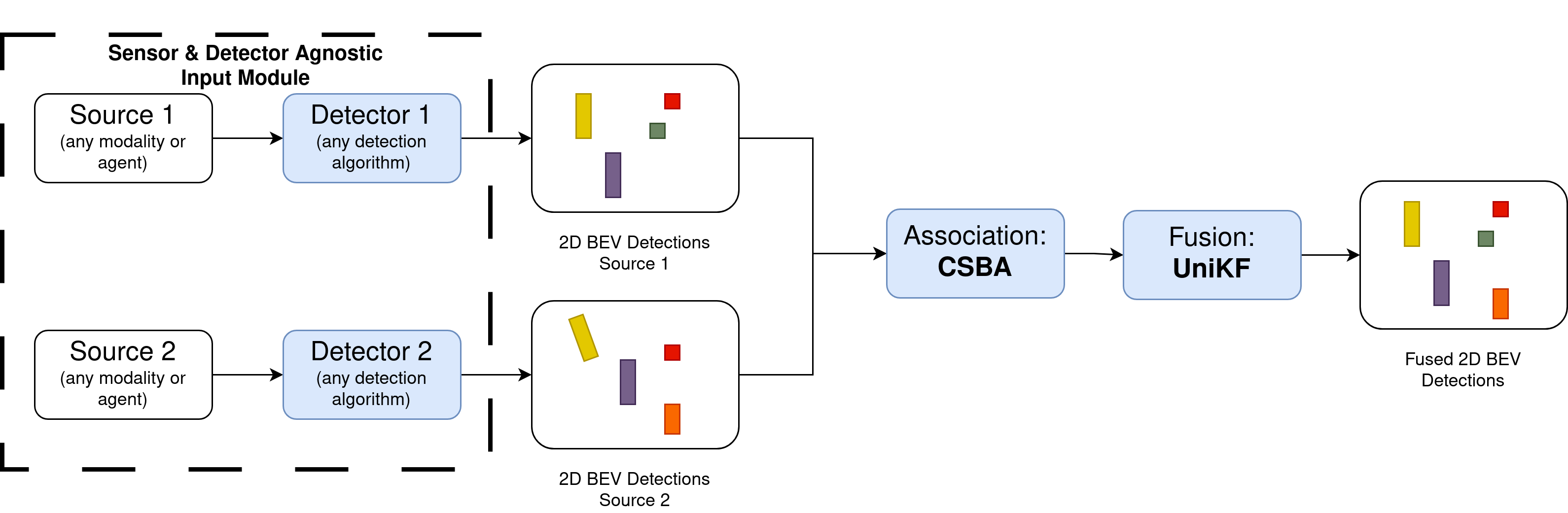}
      \caption{Overview of the proposed modular late fusion pipeline. The system is agnostic to both sensor modality and detection algorithm, supporting arbitrary input sources. Independent 2D BEV detections from each source are processed through a robust association step \textit{CSBA}, followed by uncertainty-aware fusion using \textit{UniKF}, resulting in accurate and consistent fused detections.}
   \label{fig:overview}
\end{figure*}

\section{Introduction}
\label{sec:introduction}

Accurate perception is fundamental for autonomous driving, especially in complex urban settings where sensor occlusions, limited range, and adverse weather degrade detection quality~\cite{han_collaborative_2023}. Collaborative perception, enabled by onboard sensors' communication and Vehicle-to-Everything (V2X) communication, enhances perception by sharing sensor data across multiple sensors or agents ~\cite{caillot_survey_2022, huang_multi-modal_2022}.~However, the choice of fusion strategy significantly impacts efficiency and scalability.

Early fusion methods require high bandwidth and strict time synchronization. Deep fusion demands access to proprietary models, which is impractical due to privacy and intellectual property restrictions. Late fusion, which operates at the object detection level, offers a scalable, bandwidth-efficient, and detector-model-agnostic alternative.~However, most late fusion techniques rely on heuristic-based association methods (e.g., IoU-based NMS~\cite{neubeck_efficient_2006}), which struggle in noisy and asynchronous scenarios.

To address these challenges, we propose in this paper :
\begin{enumerate}
    \item \textbf{Unified Kalman Fusion} (\textbf{UniKF}), an uncertainty-aware late fusion framework that explicitly incorporates detection uncertainties and temporal misalignment using Bayesian filtering.
    \item A noise-based evaluation protocol designed to isolate fusion performance from detection quality, enabling a targeted assessment.
    \item Extensive experiments showing that our method significantly reduces localization, orientation and size estimation errors across varying noise levels, while outperforming state-of-the-art late fusion techniques and maintaining robustness under uncertainty.
\end{enumerate}

The paper is structured as follows: \cref{sec:relatedworks} reviews the literature and positions our work within the field. \cref{sec:prob_form} defines the fusion problem and introduces our noise-injection framework for systematic evaluation. \cref{sec:propmethod} presents the proposed \textit{UniKF} pipeline, detailing its association and fusion strategies. \cref{sec:expe} presents our experiment setup and \cref{sec:results} evaluates \textit{UniKF} under various noise conditions and compares it to baseline methods. Finally, \cref{sec:Conclusion} summarizes key findings and outlines future research directions.

\section{Related Works}
\label{sec:relatedworks}
\raggedbottom

\noindent \textbf{Multi-Sensor and Multi-Agent Fusion in Autonomous Driving.}
Accurate perception is crucial for autonomous vehicles operating in complex urban environments. To improve coverage and robustness, multi-sensor or multi-agent fusion combines information from diverse modalities (e.g., LiDAR, radar, camera) or from multiple agents (e.g. vehicles, infrastructure). Traditional approaches can be divided into early fusion~\cite{zhao_coopre_2024, chen_co3_2022}, which merges raw sensor data before detection, intermediate fusion~\cite{su_collaborative_2024, liu_when2com_2020,li_v2x-dgw_2024,qiao_cobevfusion_2023, zimmer_tumtraf_2024}, where learned feature maps are shared and fused, and late fusion, which fuses only high-level bounding box detections. 

While intermediate fusion is often considered a good balance between detection accuracy and bandwidth consumption, recent studies~\cite{abdali2025} suggest that late fusion can be more efficient for both metrics. With a significantly lower median communication volume per frame (~\(2^9\) bytes compared to \(2^{19}\) bytes for deep fusion), late fusion is both scalable and bandwidth-efficient, making it well-suited for real-world multi-source perception systems. Moreover, it avoids reliance on proprietary model parameters, enhancing interoperability across different perception frameworks. 

Traditional bounding box late fusion frequently relies on basic averaging or weighted averaging of box parameters once boxes have been matched (by IoU overlap, geometric proximity, etc.). For instance, many works adapt standard NMS~\cite{neubeck_efficient_2006} to multi-sensor scenarios~\cite{bodla_soft-nms_2017,xu_model-agnostic_2023,shen_competitive_2021}, or use distance-based association~\cite{zimmer_infradet3d_2023,yu_dair-v2x_2022,picard_decentralized}. While straightforward, these heuristics can be sensitive to detection quality and may neglect the underlying uncertainties in each bounding box estimate.

Furthermore, existing surveys on multi-sensor 3D detection~\cite{wang_survey_2023,wang_multi-sensor_2024} highlight the evolution of late sensor fusion strategies, from early and feature-level approaches~\cite{MV3D2017,avod_2018} to more recent methods like CLOCs~\cite{pang2020clocs,pang2022}, which perform late fusion by combining bounding box detections. However, these late fusion techniques often rely on deep learning models, making inference computationally expensive, and primarily focus on fusing 2D image-plane bounding boxes with 3D detections.

\noindent \textbf{Bird’s Eye View Fusion.} simplifies geometric alignment across sensors by projecting data into a common top-down plane~\cite{li_delving_2023}. This approach mitigates issues related to sensor pitch, roll, and camera perspective, enabling a consistent representation of bounding box parameters (center, dimension, orientation). As a result, BEV-based methods enhance the reliability of multi-view object association and fusion.\\
\noindent \textbf{Information Fusion and Uncertainty-Awareness.}
Bayesian filtering methods, such as the Kalman Filter (KF), Extended/Unscented KF, and Particle Filters, have been widely used in information fusion, leveraging noise covariances to enhance state estimation. Broader approaches, including evidential reasoning and fuzzy logic, have also been explored~\cite{castanedo_review_2013}, demonstrating how uncertainty-aware techniques systematically improve fused estimates. Recent studies~\cite{xu_model-agnostic_2023, su_uncertainty_2023, mun_uncertainty_2023, DMSTrack} incorporate uncertainty modeling at the bounding-box level, though primarily as an auxiliary step within larger 3D detection or tracking frameworks. 

To enable real-time fusion while accounting for detection uncertainties in collaborative perception application, \cite{fadili2025} introduced an association method using only minimal object-level information. They also propose, a fusion framework combining weighted least squares with Kalman filter-based tracking. However, this approach does not handle heavily asynchronous data and latency.\\
\noindent \textbf{Challenges in Isolating Fusion Performance.}  
Most late-fusion pipelines rely on upstream detector outputs, making it difficult to separate fusion errors from those introduced by the detector itself. Minor detection biases, such as bounding box drift, can obscure the strengths or weaknesses of a fusion algorithm~\cite{harimohan2019, li_delving_2023}. As a result, fusion is often evaluated jointly with detection or tracking performance, limiting the ability to assess its independent contribution.\\
Moreover, fusion methods in autonomous driving are often evaluated on standard datasets (e.g., \textit{KITTI}~\cite{geiger_vision_2013}, \textit{nuScenes}~\cite{caesar_nuscenes_2020}, \textit{DAIR-V2X}~\cite{yu2022dairv2x}, \textit{TUMTraf}~\cite{zimmer_tumtraf_2024}) using classical metrics unfit to assess fusion task performance alone.\\
\noindent \textbf{Positioning of Our Work.}
While existing multi-agent and multi-sensor approaches often overlook systematic noise handling, our work first establishes a comprehensive \emph{Multi-Source Fusion} setup (\cref{sec:prob_form}) that injects controlled noise into ground-truth bounding-boxes. This noise-based evaluation isolates the fusion stage, letting us rigorously measure performance independent of upstream detection quality. Building on this framework, we propose \emph{Unified Kalman Fusion} (\textit{UniKF}), an uncertainty-aware late-fusion pipeline (\cref{sec:propmethod}) that leverages Kalman filtering in the BEV plane. Designed to robustly handle synchronous, asynchronous, and delayed observations, \textit{UniKF} explicitly models uncertainty throughout the fusion process, outperforming prior methods that do not systematically address the impact of noise on fusion accuracy. 

This complementary setup and method together provide a comprehensive examination of late-fusion robustness, filling a critical gap in existing multi-sensor and multi-agent fusion literature.

\section{Multi-Source Fusion: Setup, Noise, and Evaluation}
\label{sec:prob_form}
\raggedbottom
\subsection{Problem Formulation}
In the following, to focus solely on the challenges of association and fusion, without the influence of categorization errors,  we assume that all provided observations are correctly classified.

We consider a multi-agent scenario where each agent (e.g., vehicle, roadside unit) shares 3D detections projected onto the Bird’s Eye View (BEV) plane. Each detection is represented by a five-dimensional vector as in \cref{eq:obs}.
\begin{equation}
\mathbf{z} = \begin{bmatrix} x & y & w & d & \theta \end{bmatrix}^T
\label{eq:obs}
\end{equation}
where \((x, y)\) is the position within a \textit{global coordinate system}, \((w, d)\) are the object's width and depth, and $\theta$ is the Euler angle representing the rotation around the $z$-axis, which is the yaw angle in the \textit{global coordinate system}.

Our objectives are \textbf{(1) Association}, to determine which detections from different sources correspond to the same physical object, and \textbf{(2) Fusion}, to combine these associated detections into a single, more accurate bounding box.

\subsection{Noise Model}
To evaluate performance under realistic conditions, we introduce controlled noise into ground-truth bounding boxes, following the noise modeling approach of~\cite{fadili2025}. Inspired by prior work on uncertainty estimation in object detection~\cite{su_uncertainty_2023, mun_uncertainty_2023, LeUncertainty2018}, we assume all noise sources follow a zero mean Gaussian distribution.

\begin{itemize}
    \item \textbf{Distance-Dependent Position and Orientation Noise}:  
    The noise standard deviations for \(x\), \(y\), and \(\theta\) increase with the object's distance from the sensor, modeled in \cref{eq:errors}.
    \begin{equation}
    \begin{split}
        \sigma_x(\mathbf{d}) &= \sigma_{x_{0}} + k_x \mathbf{d}, \quad
        \sigma_y(\mathbf{d}) = \sigma_{y_{0}} + k_y \mathbf{d}, \\
        &\quad\quad \sigma_\theta(\mathbf{d}) = \sigma_{\theta_{0}} + k_\theta \mathbf{d}
    \end{split}
    \label{eq:errors}
\end{equation}
    where:
    \begin{itemize}
        \item \( \mathbf{d} = \sqrt{(x - x_s)^2 + (y - y_s)^2} \) is the Euclidean distance between the object and the sensor at location \((x_s, y_s)\),
        \item \(\sigma_{x_{0}}, \sigma_{y_{0}}, \sigma_{\theta_{0}}\) are the \emph{base noise levels}, representing the minimum measurement error at the sensor’s location,
        \item \(k_x, k_y, k_\theta\) are the \emph{noise increase rates}, determining how uncertainty grows with distance.
    \end{itemize}
    The noise is sampled from \cref{eq:distrib} and applied to obtain the perturbed position and orientation in \cref{eq:samp}.
    \begin{equation}
    \begin{split}
        \Delta x &\sim \mathcal{N}(0, \sigma_x(\mathbf{d})^2), \quad
        \Delta y \sim \mathcal{N}(0, \sigma_y(\mathbf{d})^2), \\
        &\quad\quad \Delta \theta \sim \mathcal{N}(0, \sigma_\theta(\mathbf{d})^2)
    \end{split}
    \label{eq:distrib}
    \end{equation}    
    \begin{equation}
    x' = x + \Delta x, \quad
    y' = y + \Delta y, \quad
    \theta' = \theta + \Delta \theta.
    \label{eq:samp}
    \end{equation}

    \item \textbf{Multiplicative Size Noise}:  
    The width and depth are perturbed using scaling factors \(\alpha\) and \(\beta\), drawn from \cref{eq:alphabeta} ensuring realistic variations in bounding box dimensions obtained in \cref{eq:alphabetaprime}.
    \begin{equation}
    \alpha \sim \mathcal{N}(1, \sigma_\alpha^2), \quad
    \beta \sim \mathcal{N}(1, \sigma_\beta^2),
    \label{eq:alphabeta}
    \end{equation}
    \begin{equation}
    w' = \alpha w, \quad d' = \beta d
    \label{eq:alphabetaprime}
    \end{equation}
    To prevent extreme values, soft-clipping~\cite{burkardt_truncated_2009} is applied, ensuring the resulting width and depth remain within valid bounds.
\end{itemize}

Although real-world applications also involve systematic calibration offsets and localization drift, we focus here on random noise to isolate the core fusion challenges. Future work will incorporate such additional biases.

\subsection{Evaluation Protocol}
We use the following steps to assess our fusion approach independently of upstream detector performance:

\begin{enumerate}
    \item \textbf{Noisy Data Generation}: For each scene, we generate multiple noisy realizations of ground-truth bounding boxes (one set per sensor and/or agent) by sampling from the defined noise model.
    \item \textbf{Association \& Fusion}: Detections are matched across agents using an association strategy, then fused using the selected fusion method.    
    \item \textbf{Evaluation Metrics}: We match predicted bounding boxes to ground truth using object identifiers, ensuring a perfect pairing. When multiple predictions correspond to the same ground truth, only the closest prediction is retained as a true positive. Detection performance is evaluated using:
    \begin{itemize}
        \item \textbf{Precision and Recall}: Standard detection metrics assessing the proportion of correctly fused objects.        
    \item \textbf{Average Translation Error (ATE)}: A variant of the nuScenes \cite{caesar_nuscenes_2020} translation error metric that accounts for both true positives and false positives (\cref{eq:ate}).
    \begin{equation}
        \text{ATE} = \frac{1}{N} \sum_{i=1}^{N} \left\| \begin{bmatrix} x_i^{\text{pred}} \\ y_i^{\text{pred}} \end{bmatrix} - \begin{bmatrix} x_i^{\text{gt}} \\ y_i^{\text{gt}} \end{bmatrix} \right\|_2
    \label{eq:ate}
    \end{equation}
    where \( (x_i^{\text{pred}}, y_i^{\text{pred}}) \) and \( (x_i^{\text{gt}}, y_i^{\text{gt}}) \) are the predicted and ground truth object centers, and \( N \) is the total number of predictions $FP+TP$.

    \item \textbf{Average Orientation Error (AOE)}: A variant of the nuScenes \cite{caesar_nuscenes_2020} orientation error metric that accounts for both true positives and false positives (\cref{eq:aoe}).
    \begin{equation}
        \text{AOE} = \frac{1}{N} \sum_{i=1}^{N} \min \left( | \theta_i^{\text{pred}} - \theta_i^{\text{gt}} |, 2\pi - | \theta_i^{\text{pred}} - \theta_i^{\text{gt}} | \right)
    \label{eq:aoe}
    \end{equation}
    where \( \theta_i^{\text{pred}} \) and \( \theta_i^{\text{gt}} \) are the predicted and ground truth yaw angles, respectively.

    \item \textbf{Average Dimension Error (ADE)}:  
    Computes the Euclidean difference in width and depth between predicted and ground truth objects (\cref{eq:ade}).
    \begin{equation}
        \text{ADE} = \frac{1}{N} \sum_{i=1}^{N} \left\| \begin{bmatrix} w_i^{\text{pred}} \\ d_i^{\text{pred}} \end{bmatrix} - \begin{bmatrix} w_i^{\text{gt}} \\ d_i^{\text{gt}} \end{bmatrix} \right\|_2
    \label{eq:ade}
    \end{equation}
    where \( w_i^{\text{pred}}, d_i^{\text{pred}} \) and \( w_i^{\text{gt}}, d_i^{\text{gt}} \) represent the width and depth of the predicted and ground truth bounding boxes, respectively.
        
    \end{itemize}
    The mean values of these metrics, denoted as \textit{mATE}, \textit{mAOE}, and \textit{mADE}, are computed by averaging \textit{ATE}, \textit{AOE}, and \textit{ADE} over all frames. Traditionally, these metrics evaluate only true positives, which disadvantage methods that produce fewer false positives.
    \item \textbf{Robustness Evaluation}: The fusion pipeline is tested over multiple trials $N$ to assess robustness, measuring performance metrics across random noise generation.
\end{enumerate}


\section{UniKF: An Uncertainty-Aware Late-Fusion Pipeline}
\label{sec:propmethod}
\raggedbottom
We introduce in Fig.~\ref{fig:overview} : \textbf{Unified Kalman Fusion (UniKF)}, a novel uncertainty-aware late-fusion pipeline designed to integrate high-level bounding boxes from multiple sources in the Bird’s Eye View (BEV). Our pipeline operates in three main steps:
\begin{enumerate}
    \item \textbf{Association}: Identify which detections from different sources refer to the same object. We employ the \emph{Combined-Score Based Association (CSBA)} approach introduced in~\cite{fadili2025} to pair detections across sources using a cost function that integrates uncertainties while combining scores for center alignment, dimension similarity, and orientation consistency.
    \item \textbf{Kalman Filter-Based Fusion}: Fuse associated detections over time with an uncertainty-aware Kalman filter.
    \item \textbf{Time-Sensitive Updates}: Adapt the filter to handle measurements arriving with various timing situations : synchronous, asynchronous, and out-of-sequence observations induced by communication delays.
\end{enumerate}

\subsection{Kalman Filter Formulation}
Each object's state is modeled as in \cref{eq:state}.
\begin{equation}
\mathbf{x} = \begin{bmatrix} x & y & v_x & v_y & w & d & \theta \end{bmatrix}^T
\label{eq:state}
\end{equation}
where $(x, y)$ represents the object's position, $(v_x, v_y)$ are the velocities along each axis in the BEV plane, $(w, d)$ are the object's width and depth, and $\theta$ denotes its orientation.

We adopt the constant velocity (CV) model, as it offers the best trade-off between accuracy and robustness in multi-agent bounding box fusion, according to~\cite{fadili2025}. Following the formulation in~\cite{baisa_derivation_2020}, the state evolution is governed by the transition matrix $\mathbf{F}$, which updates the position $(x, y)$ based on velocity $(v_x, v_y)$ over a time step $\Delta t$, while keeping other state components unchanged (\cref{eq:transition}).
\begin{equation}
    \mathbf{x}_{t+1} = \mathbf{F} \mathbf{x}_t + \mathbf{q}_t.
\label{eq:transition}
\end{equation}
Here, $\mathbf{q}_t$ represents Gaussian process noise, modeling uncertainties in state transitions.

A measurement $\mathbf{z}_t$ from any agent at time step \(t\) is denoted as in \cref{eq:obs}.
The measurement model relates the observation \(\mathbf{z}_t\) to the true state \(\mathbf{x}_t\) through \cref{eq:meas}.
\begin{equation}
    \mathbf{z}_t = \mathbf{H} \mathbf{x}_t + \mathbf{r}_t,
\label{eq:meas}
\end{equation}
where $\mathbf{H}$ is the measurement matrix that extracts the observed components from the state vector, and $\mathbf{r}_t$ is the measurement noise, assumed to be Gaussian with covariance $\mathbf{R_{t}}$.~The Kalman filter updates its state using standard Kalman equations, incorporating these noisy observations to refine the state estimate and its covariance matrix $\mathbf{P_{t}}$.

\subsection{Time-Sensitive Fusion}
\label{sec:time_sensitive_fusion}
\begin{figure}[h]
  \centering  
   \includegraphics[width=1\linewidth]{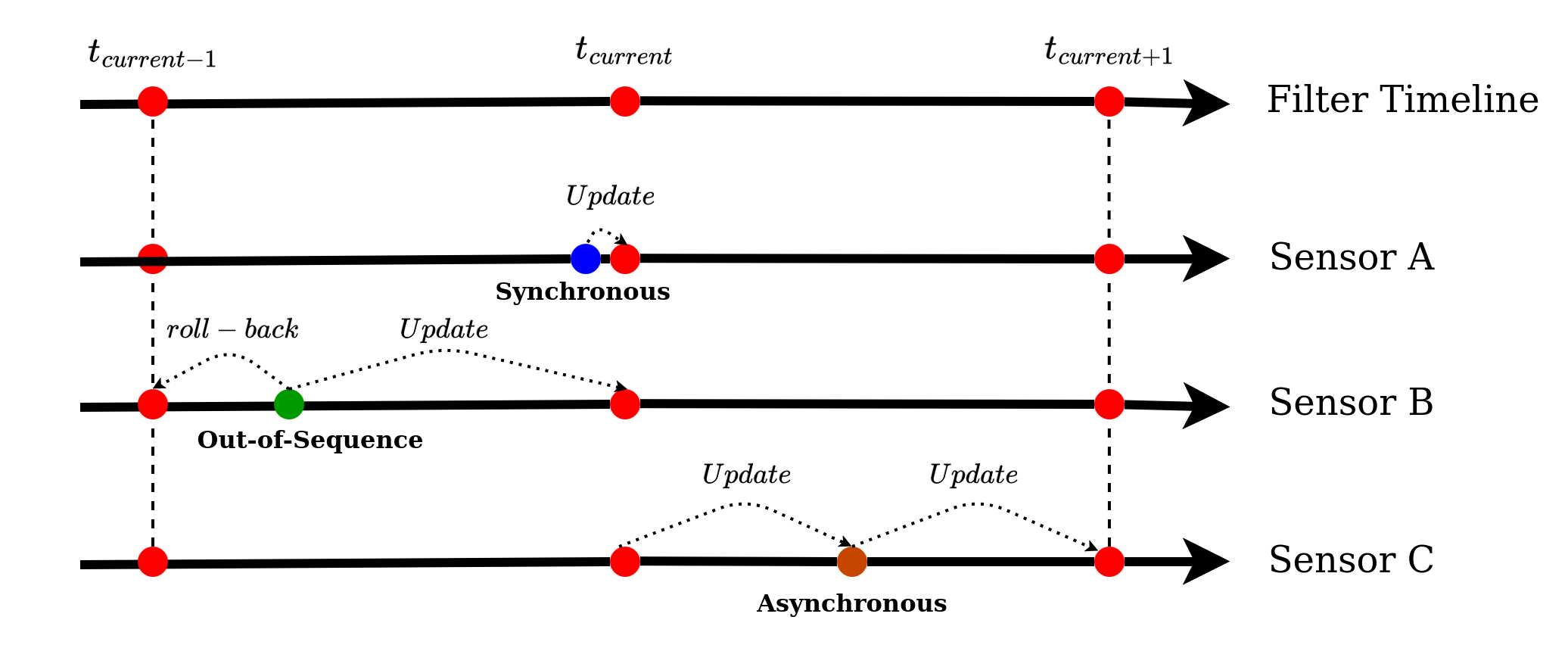}
      \caption{\textit{UniKF} time consistency handling for different sensor types. Sensor A provides synchronous data updates within a predefined tolerance window, allowing direct state updates. Sensor B delivers delayed measurements; thus, the filter rolls back to the nearest historical state, incorporates the delayed measurement, and propagates forward to the current timestamp. Sensor C provides asynchronous data at irregular intervals, triggering immediate filter updates upon data arrival. Consequently, the proposed \textit{UniKF} dynamically combines time-based and event-triggered update mechanisms.}
   \label{fig:kalman_time}
\end{figure}

A key feature of our method \textit{UniKF} is its capacity to integrate measurements arriving at different times. Fig. \ref{fig:kalman_time} illustrates \textit{UniKF}'s approach to managing data synchronization and latency challenges effectively. We distinguish three scenarios:

\begin{itemize}
    \item \textbf{Synchronous Measurements}:
    These have timestamps closely matching the filter’s current time \(t_{\mathrm{current}}\). If the absolute difference 
    \(\lvert t_{\mathrm{meas}} - t_{\mathrm{current}}\rvert\) is within a small tolerance $\varepsilon_{s}$, the system simply applies a normal Kalman update at \(t_{\mathrm{current}}\).

    \item \textbf{Out-of-Sequence Measurements}:  
    These arrive with timestamps \emph{earlier} than the filter’s current time, i.e.,  
    \( t_{\mathrm{meas}} < t_{\mathrm{current}} - \varepsilon_{s} \). To incorporate delayed data, the filter first retrieves a stored state from a history buffer corresponding to a time at or just before \( t_{\mathrm{meas}} \). It then reverts to this earlier time and state—a process often called \emph{roll back}. The delayed measurement is used to perform a standard Kalman update, after which the filter is forward-propagated to the current time.
    
    This process acts like a smoothing mechanism, ensuring delayed but potentially valuable observations are integrated without violating time consistency. 
    
    \item \textbf{Asynchronous Observations}: 
    These are sensor measurements that have timestamps beyond the filter’s current timeline, relative to the primary sensor, i.e., 
    \(t_{\mathrm{meas}} > t_{\mathrm{current}} + \varepsilon_{s}\). However, these are not predictions of an unobserved future, as in trajectory forecasting or intent prediction tasks. Instead, they result from fusing sensors operating at different frequencies, where an observation from a secondary sensor falls ahead of the primary sensor’s present state. To correctly integrate such measurements, we propagate the filter state forward to \(t_{\mathrm{meas}}\), apply the measurement update, and then re-propagate to the primary sensor's timeline. This ensures that an observation is fused as soon as it arrives, without waiting for filter's next iteration, which could be beneficial for example when the current estimate has high uncertainties or when an object's dynamic varies suddenly.
    \end{itemize}

We also define a maximum latency threshold \(\Delta_{\mathrm{max}}\). If an incoming measurement is too far in the past or future relative to \(t_{\mathrm{current}}\), we choose to discard it. 

\subsection{Uncertainty-Aware Fusion Process}
\begin{enumerate}
    \item \textbf{Initialization}:
    For a newly observed object, initialize the Kalman filter state with the first detection (setting velocity to zero or a small prior). The initial covariance $\mathbf{P_{t}}$ captures the uncertainty in position, dimensions, and orientation based on the reported detection accuracy.
    \item \textbf{Subsequent Updates}:
    Each time a new measurement arrives (synchronous or otherwise), the filter:
    \begin{itemize}
        \item Reverts or advances the timeline if needed (as described in \cref{sec:time_sensitive_fusion}).
        \item Performs a standard Kalman update, where the measurement covariance $\mathbf{R_{t}}$ is set according to the associated bounding box noise level.
    \end{itemize}
    \item \textbf{Extracting the Fused Box}:
    The final 7D state $\mathbf{x}$ is interpreted as the object’s fused position, velocity, size, and orientation.
\end{enumerate}

\subsection{Limitations}
\label{subsec:lim}
Our method utilizes a constant-velocity assumption, which may be insufficient for highly maneuvering objects. Additionally, we do not explicitly model systematic calibration offsets or large clock drifts; hence, extreme misalignments might degrade performance. Nonetheless, by focusing on Gaussian noise and timing irregularities, we highlight the robustness and adaptability of this approach for typical multi-sensor or multi-agent scenarios.

Furthermore, although our \textit{UniKF} method naturally delivers tracking results by maintaining object states over time, in this work, we have deliberately chosen to prioritize fusion performance before evaluating tracking behavior. This decision is motivated by two factors: (1) our primary objective is to first validate the accuracy and robustness of the fusion process before analyzing the full impact on tracking performance, and (2) due to space constraints, we could not integrate all tracking-related evaluation metrics in this paper. 

Beyond tracking considerations, our study does not include comparisons with early or deep fusion methods, as these typically rely on sharing raw sensor data or intermediate feature representations—requirements that extend beyond the scope of this work. 

\section{Experiments}
\label{sec:expe}
\raggedbottom

\noindent \textbf{Dataset.} We evaluate our approach on the complete nuScenes validation dataset~\cite{caesar_nuscenes_2020}, which consists of 150 scenes and approximately 145,000 annotated objects, covering diverse urban driving scenarios with multiple object types, including cars, buses, trucks, pedestrians, bicycles, motorcycles and movable objects such as barrier etc. 

Unlike datasets that focus primarily on vehicle detection~\cite{xu2022opv2v, hu2022v2xset} in simulated environments~\cite{li2021v2xsim}, constrained scenarios such as highways and intersections~\cite{zimmer_tumtraf_2024}, or those with restricted access~\cite{yu2022dairv2x}, nuScenes provides an open-source, diverse urban environment. It incorporates Vulnerable Road Users (VRUs) and various static obstacles, making it particularly well-suited for evaluating fusion performance across a wide range of object dynamics and interaction complexities. Nonetheless, our method is inherently \textbf{generalizable} to other datasets, as it does not rely on dataset-specific learning. Instead, it remains \textbf{detector-agnostic} and \textbf{dataset-agnostic}.

To simulate multi-source detection scenarios, we adopt the approach of~\cite{fadili2025}, applying controlled Gaussian noise to ground-truth bounding boxes to create synthetic imperfect detections. We assume a dual-sensor/agent setup where one sensor is at the ego vehicle position, while the other is placed randomly in each frame. The base noise levels are empirically set with equal position noise parameters, $\sigma_{x_{0}} = \sigma_{y_{0}}$, and noise increase rate $k_{x} = k_{y} = 0.01$. Orientation noise is modeled with a fixed value of $k_{\theta} = 0.1$, while width and depth uncertainties are assigned the same noise levels, $\sigma_{\alpha} = \sigma_{\beta}$.\\
\noindent \textbf{Noise Parameterization.} The choice of noise levels in Tab. \ref{tab:noise_levels} is guided by the performance of state-of-the-art 3D object detectors on the nuScenes benchmark, particularly their translation, orientation, and scale (1-IoU) errors. LiDAR-based detectors generally achieve position errors between $0.2m$ and $0.5m$, while camera-based methods can exceed $1.0m$, leading us to define two noise levels: low ($0.2m$), and moderate ($0.5m$). Orientation errors vary significantly, with vision-based detectors achieving an error of $0.2^\circ$, whereas some LiDAR and Radar approaches may exceed $1^\circ$, justifying our choice of $\sigma_{\theta_{0}} = {0.2^\circ, 5^\circ}$. Scale estimation errors range from $0.22$ to $0.61$, so we set dimension noise levels at $\sigma_{\alpha} = {0.2, 0.5}$. The high-noise configuration \textit{"Noise 3"} is designed to reflect the uncertainty levels associated with less accurate object detectors, particularly those whose performance metrics may be biased due to training, validation, and testing being conducted on the same dataset~\cite{gawlikowski_2023}. This setup ensures a more realistic uncertainties that better approximate real-world deployment scenarios.
\begin{table}[h]
    \caption{Noise configurations for position, yaw, and size perturbations applied to ground truth annotations from nuScenes. The IOU and distance threshold used in baselines are presented here.}
    \centering
    \label{tab:noise_levels}
    \begin{tabular}{|l|c|c|c|c|c|}
        \hline
        \textbf{Noise Level} & \textbf{$\sigma_{x_{0}}$} & \textbf{$\sigma_{\theta_{0}}$} & \textbf{$\sigma_{\alpha}$} & \textbf{IOU th} & \textbf{Dist th}  \\
        \hline
        \textbf{Noise 1}      & $0.2m$  & $0.2^\circ$   & $0.2$ & $0.5$ & $3m$ \\
        \textbf{Noise 2}   & $0.5m$  & $5.0^\circ$  & $0.5$ & $0.5$ & $3m$  \\
        \textbf{Noise 3}      & $1.0m$  & $10^\circ$  & $1.0$ &  $0.3$ & $3m$  \\
        \hline
    \end{tabular}    
\end{table}\\
\noindent \textbf{CSBA parameters.} For the association method, we use exactly the same parameters setting than authors of~\cite{fadili2025}.\\
\noindent \textbf{Fusion Parameters.} We consider time tolerance $\varepsilon_{s}~=~10ms$; latency threshold $\Delta_{\mathrm{max}}~=~500ms$ and number of trials $N~=~5$.\\
\noindent \textbf{Baseline Methods.}  
We compare our \textit{UniKF} framework against state-of-the-art late fusion approaches, including classical Non-Maximum Suppression (\textit{NMS-STD})~\cite{neubeck_efficient_2006} —used in approaches such as DMSTrack\cite{DMSTrack}, AB3DMOT~\cite{weng_ab3dmot_2020}, and V2V4Real~\cite{yang2023v2v4real}—, Promote-Suppress Aggregation (\textit{PSA})~\cite{xu_model-agnostic_2023}, and Weighted Box Fusion (\textit{WBF})~\cite{shen_competitive_2021}, all of which merge detections using IoU matching. Additionally, we evaluate \textit{NMS-GIoU}~\cite{giou2019}, which refines suppression decisions using Generalized IoU. 

We further compare against distance-based late fusion methods, including \textit{InfraDet3D-Late}~\cite{zimmer_infradet3d_2023} and \textit{DAIR-V2X-Late}~\cite{yu2022dairv2x}, both adapted to handle 3D detections projected onto the BEV plane. Finally, we include \textit{CSBA+WLS}~\cite{fadili2025}, which combines \textit{CSBA} association with Weighted Least Squares (\textit{WLS}) fusion. While \textit{CSBA+WLS} does not explicitly handle asynchronous or latent data, it mitigates misalignment using a temporal sliding window set to $100ms$ to approximate synchrony.

The thresholds used for IOU-based and distance-based association are presented in Tab. \ref{tab:noise_levels}, ensuring consistency with the original configurations defined by each method’s authors.

\textit{We do not compare against early and deep fusion methods} as discussed in \cref{subsec:lim}.  Additionally, no AI-based late fusion method for BEV object fusion exists in the current literature of collaborative perception and sensor fusion for autonomous driving, and thus no direct comparison is possible.

\section{Results and Discussions}
\label{sec:results}
\raggedbottom
This section presents a comprehensive evaluation of our \textit{CSBA+UniKF} in comparison to existing late fusion methods under varying noise conditions. 

\subsection{Comparison of Baseline and Proposed Method Performance Under Varying Noise Levels}
\label{subsec:comp_perf_glob}

\begin{table*}[t]
\caption{Comparison of late fusion methods under three different noise levels.}
\centering
\scriptsize
\label{tab:performance_metrics_all_dataset}
\begin{tabular}{lllccccc}
\toprule
\textbf{Detector 1} & \textbf{Detector 2} & \textbf{Method} & \textbf{mATE (m)} & \textbf{mADE (m)} & \textbf{mAOE (deg)} & \textbf{Precision} & \textbf{Recall} \\
\midrule
\multirow{7}{*}{Noise 1} & \multirow{7}{*}{Noise 1} 
    & NMS-STD \cite{neubeck_efficient_2006} & $0.72\pm0.18$ & $0.79\pm0.31$ & $3.14\pm1.18$ & $\mathbf{100}$ & $\mathbf{100}$ \\
    & & PSA \cite{xu_model-agnostic_2023} & $1.44\pm0.31$ & $1.58\pm0.58$ & $6.28\pm1.93$ & $50.0$ & $\mathbf{99.9}$ \\
    & & WBF \cite{shen_competitive_2021} & $1.29\pm0.32$ & $1.29\pm0.49$ & $5.57\pm1.89$ & $57.0$ & $99.0$ \\
    & & NMS-GIoU \cite{giou2019} & $1.33 \pm 0.32$ & $1.39 \pm 0.51$ & $5.74 \pm 1.95$ & $56.0$ & $\mathbf{99.9}$ \\
    & & InfraDet3D-Late~\cite{zimmer_infradet3d_2023} & $0.75\pm0.23$ & $0.59\pm0.24$ & $3.59\pm2.74$ & $98.0$ & $\mathbf{100}$ \\
    & & DAIR-V2X-Late~\cite{yu2022dairv2x} & $0.58 \pm 0.21$ & $0.59 \pm 0.24$ & $2.86 \pm 1.89$ & $98.0$ & $\mathbf{100}$ \\
    & & CSBA+WLS~\cite{fadili2025} & $0.53\pm0.15$ & $0.57\pm0.23$ & $2.44\pm1.32$ & $\mathbf{99.9}$ & $\mathbf{100}$ \\
    & & \textbf{CSBA+UniKF (Ours)} & $\mathbf{0.51\pm0.12}$ & $\mathbf{0.56\pm0.22}$ & $\mathbf{2.35\pm1.34}$ & $99.5$ & $\mathbf{100}$ \\

\midrule
\multirow{7}{*}{Noise 2} & \multirow{7}{*}{Noise 2}  
    & NMS-STD \cite{neubeck_efficient_2006} & $2.20 \pm 0.37$ & $3.66 \pm 1.33$ & $13.96 \pm 2.89$ & $50.0$ & $\mathbf{100}$ \\
    & & PSA \cite{xu_model-agnostic_2023} & $2.20\pm0.37$ & $3.65\pm1.34$ & $13.97\pm2.87$ & $50.0$ & $\mathbf{100}$ \\
    & & WBF \cite{shen_competitive_2021} & $2.16\pm0.37$ & $3.56\pm1.33$ & $13.70\pm2.87$ & $51.0$ & $\mathbf{100}$ \\
    & & NMS-GIoU \cite{giou2019} & $2.16 \pm 0.38$ & $3.56 \pm 1.31$ & $13.66 \pm 2.81$ & $51.1$ & $99.9$ \\
    & & InfraDet3D-Late~\cite{zimmer_infradet3d_2023} & $1.26\pm0.35$ & $1.52\pm0.68$ & $8.28\pm3.67$ & $93.0$ & $\mathbf{100}$ \\
    & & DAIR-V2X-Late~\cite{yu2022dairv2x} & $1.04 \pm 0.34$ & $1.51 \pm 0.68$ & $6.67 \pm 3.01$ & $92.7$ & $\mathbf{100}$ \\
    & & CSBA+WLS~\cite{fadili2025}& $0.81\pm0.17$ & $\mathbf{1.31\pm0.53}$ & $5.29\pm2.07$ & $\mathbf{99.9}$ & $\mathbf{100}$ \\
    & & \textbf{CSBA+UniKF (Ours)} & $\mathbf{0.80\pm0.18}$ & $1.32\pm0.53$ & $\mathbf{5.28\pm2.10}$ & $\mathbf{99.9}$ & $\mathbf{100}$ \\
\midrule
\multirow{7}{*}{Noise 3} & \multirow{7}{*}{Noise 3}  
    & NMS-STD \cite{neubeck_efficient_2006} &$3.44 \pm 0.48$ & $6.54 \pm 2.46$ & $21.99 \pm 3.92$ & $50.0$ & $\mathbf{100}$ \\
    & & PSA \cite{xu_model-agnostic_2023} & $3.45\pm0.49$ & $6.55\pm2.43$ & $21.96\pm3.94$ & $50.0$ & $\mathbf{100}$ \\
    & & WBF \cite{shen_competitive_2021} & $3.25\pm0.51$ & $6.03\pm2.34$ & $20.61\pm3.88$ & $54.0$ & $99.7$ \\
    & & NMS-GIoU \cite{giou2019} & $3.33 \pm 0.50$ & $6.17 \pm 2.29$ & $20.99 \pm 3.96$ & $52.4$ & $\mathbf{99.9}$ \\
    & & InfraDet3D-Late~\cite{zimmer_infradet3d_2023} & $2.41\pm0.55$ & $3.77\pm1.61$ & $15.19\pm4.89$ & $78.0$ & $\mathbf{100}$ \\
    & & DAIR-V2X-Late~\cite{yu2022dairv2x} & $2.24 \pm 0.58$ & $3.77 \pm 1.62$ & $13.29 \pm 4.28$ & $77.8$ & $\mathbf{100}$ \\
    & & CSBA+WLS~\cite{fadili2025} & $1.30\pm0.24$ & $\mathbf{2.55\pm1.01}$ & $8.54\pm3.19$ & $\mathbf{100}$ & $\mathbf{100}$ \\
    & & \textbf{CSBA+UniKF (Ours)} & $\mathbf{1.29\pm0.23}$ & $\mathbf{2.55\pm1.02}$ & $\mathbf{8.53\pm3.10}$ & $\mathbf{100}$ & $\mathbf{100}$ \\

\midrule
\multirow{7}{*}{Noise 1} & \multirow{7}{*}{Noise 3} 
    & NMS-STD \cite{neubeck_efficient_2006} & $2.17 \pm 0.45$ & $1.32 \pm 0.45$ & $11.99 \pm 3.29$ & $58.55$ & $99.87$ \\
    & & PSA \cite{xu_model-agnostic_2023} & $2.44 \pm 0.40$ & $1.60 \pm 0.56$ & $14.10 \pm 3.10$ & $50.02$ & $\mathbf{100}$ \\

    & & WBF \cite{shen_competitive_2021} & $2.14 \pm 0.45$ & $1.23 \pm 0.43$ & $11.99 \pm 3.24$ & $58.70$ & $99.79$ \\
    & & NMS-GIoU \cite{giou2019} & $2.24 \pm 0.44$ & $1.38 \pm 0.46$ & $12.50 \pm 3.21$ & $56.27$ & $99.87$ \\
    & & InfraDet3D-Late ~\cite{zimmer_infradet3d_2023} & $1.34 \pm 0.41$ & $0.73 \pm 0.30$ & $8.12 \pm 3.21$ & $87.39$ & $\mathbf{100}$ \\
    & & DAIR-V2X-Late~\cite{yu2022dairv2x} & $1.33 \pm 0.41$ & $\mathbf{0.74 \pm 0.31}$ & $8.13 \pm 3.26$ & $87.41$ & $\mathbf{100}$ \\
    & & CSBA+WLS~\cite{fadili2025} & $\mathbf{0.67 \pm 0.15}$ & $0.76 \pm 0.30$ & $4.40 \pm 1.41$ & $\mathbf{100}$ & $\mathbf{100}$ \\
    & & \textbf{CSBA+UniKF (Ours)} & $\mathbf{0.67\pm0.15}$ & $0.76\pm0.30$ & $\mathbf{4.38\pm1.43}$ & $\mathbf{100}$ & $\mathbf{100}$ \\

\bottomrule
\end{tabular}
\end{table*}

We present a comparison of state-of-the-art late fusion methods with our proposed approach across different noise configurations. 

The results demonstrate that \textit{CSBA+UniKF} consistently outperforms existing late fusion methods across all noise configurations, achieving lower translation, dimension, and orientation errors while maintaining high precision and recall. \textit{CSBA+WLS} exhibits comparable performance in position and dimension estimation across all noise levels but shows a slight increase in orientation error (up to $0.09^\circ$ higher than \textit{CSBA+UniKF}). 

This result arises because \textit{nuScenes} exhibits relatively low asynchronicity and latency issues, with observed delays ranging between $1 ms$ and $100 ms$. In such cases, a sliding temporal window of $100 ms$, as used in \textit{CSBA+WLS}, effectively mitigates minor time misalignments. Additionally, since \textit{CSBA+WLS} performs a purely observation-driven fusion without incorporating prior state uncertainty, it can sometimes yield equivalent or better results in synchronous settings, particularly when the motion model in \textit{UniKF} does not provide significant additional benefits (e.g. static objects).

However, in more challenging scenarios with higher latency or asynchronous detections, \textit{UniKF} demonstrates greater robustness by leveraging prior estimates to refine fusion outcomes, ensuring a more stable and accurate estimation under varying uncertainty conditions.

Under \textit{Noise 1}, both \textit{NMS-STD} and \textit{DAIR-V2X-Late} achieve relatively good performance, with the latter yielding a translation error of $0.58m$ and an orientation error of $2.44^\circ$. However, as the noise level increases (\textit{Noise 3}) and in mixed scenarios combining \textit{Noise 1} and \textit{Noise 3}, all errors increase significantly, and the performance gap between \textit{WLS}, \textit{UniKF}, and the other methods widen. Moreover, error variability becomes more pronounced, and precision deteriorates. This degradation is primarily due to the limitations of IoU and distance-based association strategies, which struggle with small objects and high-noise conditions, often rejecting valid associations when detections exceed predefined fixed thresholds.

\textit{Precision} values also highlight a key distinction between methods: approaches like \textit{PSA} and \textit{WBF} show a precision of around $50\%$, which suggests that no fusion is effectively performed since the number of false positives equals the number of true positives. In contrast, \textit{CSBA+UniKF} achieves near-perfect precision ($99.5\%$–$100\%$), demonstrating its ability to correctly associate detections and minimize false positives.

\subsection{Comparison of Translation and Orientation Errors Across Object Categories}
\label{subsec:comp_per_class}

\begin{figure*}[h]
  \centering  
   \includegraphics[width=0.97\linewidth]{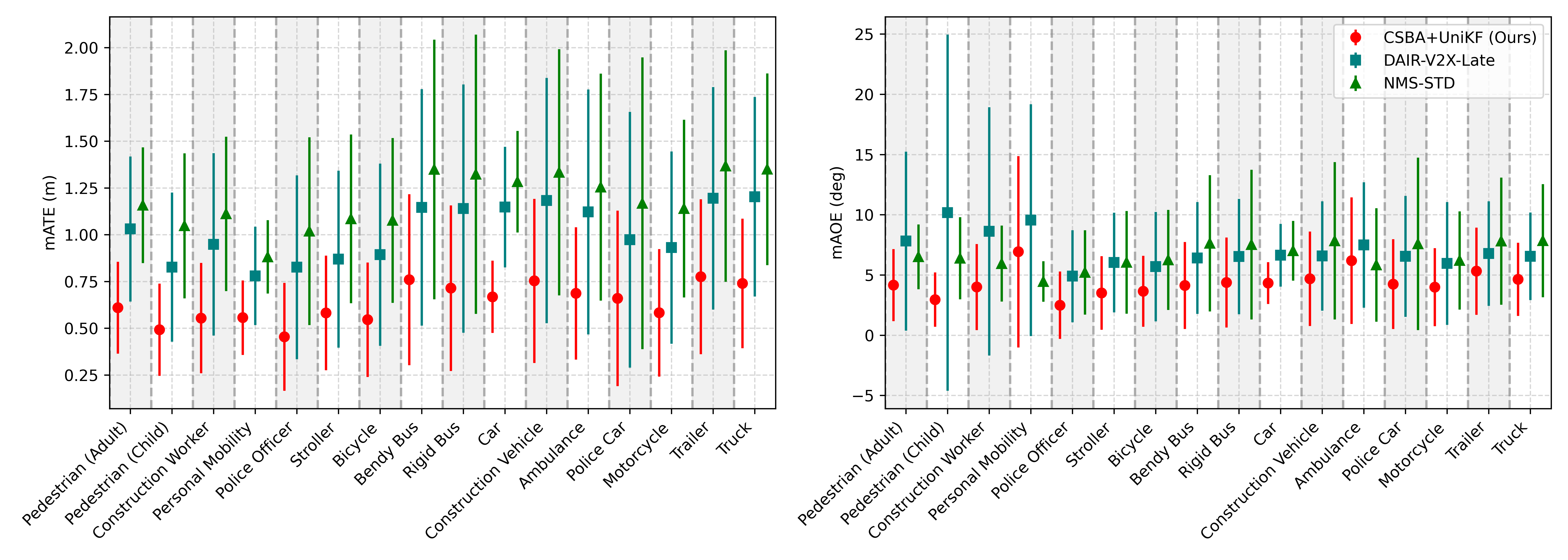}
    \caption{Per-class comparison of \textit{mATE} and \textit{mAOE} for our method \textit{CSBA+UniKF}, \textit{DAIR-V2X-Late}~\cite{yu2022dairv2x}, and \textit{NMS-STD}~\cite{neubeck_efficient_2006} under the combined \textit{Noise 1} and \textit{Noise 3} configuration.}
   \label{fig:perobject}
\end{figure*}

Fig.~\ref{fig:perobject} presents a per-class comparison of \textit{mATE} and \textit{mAOE} across \textit{CSBA+UniKF}, \textit{DAIR-V2X-Late}, and \textit{NMS-STD} under the combined \textit{Noise 1} and \textit{Noise 3} configuration. The results demonstrate that \textit{CSBA+UniKF} (red circles) consistently achieves lower errors across most object categories, reinforcing its robustness in heterogeneous traffic scenarios.

In pedestrian-related classes, such as \textit{Adult} and \textit{Construction Worker}, \textit{CSBA+UniKF} outperforms baseline methods by reducing translation and orientation errors while maintaining greater stability, which is crucial for vulnerable road users detection. For vehicle categories, including \textit{Car} and \textit{Truck}, it maintains competitive performance with consistently lower errors. While \textit{DAIR-V2X-Late} performs well in some vehicle classes, it exhibits higher variability in orientation errors, whereas \textit{NMS-STD} struggles with precise localization, showing higher translation errors and greater error variance.

Overall, the results confirm that \textit{CSBA+UniKF} achieves a trade-off between translation and orientation accuracy while maintaining stability across diverse object categories. 

\subsection{Evaluating Association Strategies in UniKF Fusion}
\label{subsec:comp_assoc}
\begin{figure}[h]
  \centering  
   \includegraphics[width=1\linewidth]{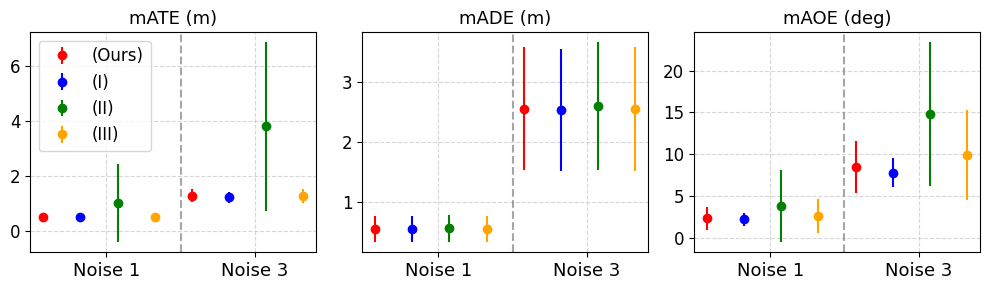}
      \caption{Impact of different association algorithms combined with \textit{UniKF} fusion. We evaluate on a setup combining two detection sources with Noise 1 configuration and two sources with Noise 3 configuration. \textbf{Ours}: \textit{UniKF} with \textit{CSBA} matching; \textbf{(I)}: \textit{UniKF} with perfect association using ground truth IDs; \textbf{(II)}: \textit{UniKF} with IoU-based association ; \textbf{(III)}: \textit{UniKF} with Euclidean distance-based association. Each point represents the metric's value, and the bars indicate the standard deviation.}
   \label{fig:comp_assoc}
\end{figure}

~\cref{fig:comp_assoc} presents the impact of different association strategies combined with \textit{UniKF} fusion under two noise configurations, \textit{Noise 1} and \textit{Noise 3}. The results highlight the effectiveness of our proposed \textit{CSBA+UniKF} method, which consistently achieves high precision ($99.5\%$–$100\%$) and recall ($100\%$) across both noise levels. Under \textit{Noise~1}, \textit{CSBA+UniKF} achieves the lowest mATE of $0.51m$ and mADE of $0.56m$. The results are comparable to those of \textit{UniKF with Ground Truth Association} (in blue), further validating the accuracy of the \textit{CSBA} association approach.

For \textit{Noise 3}, \textit{CSBA+UniKF} maintains robust performance with an mATE of $1.29m$, outperforming the \textit{IoU-based} and \textit{Euclidean distance-based} association methods. Notably, \textit{IoU-based association} exhibits the highest mATE and mAOE, indicating its sensitivity to high level noise. Similarly, the \textit{Euclidean distance-based} association shows a higher orientation error ($9.93^\circ$) than our proposed method, due to the limitations of fixed-threshold distance.

These findings emphasize that our \textit{CSBA+UniKF} achieves ground truth comparable mean performance with good stability. In contrast, the high standard deviation in IoU-based demonstrates that improper association strategies introduce inconsistency, making them unreliable for real-world deployment. 
\subsection{Comparison of Single Detector and Our Collaborative Fusion Performance}
\label{subsec:comp_collab}

\begin{table}[h]
\caption{Comparison of single-detector perception with our collaborative method using CSBA association followed by UniKF fusion across different noise settings. N1, N2, and N3 stand respectively for Noise 1, Noise 2, and Noise 3}
\centering
\scriptsize
\begin{tabular}{llccc}
\toprule
\textbf{Method} & \textbf{mATE (m)} & \textbf{mADE (m)} & \textbf{mAOE ($^\circ$)} \\
\midrule
    \textbf{CSBA+UniKF} (N1) & $\mathbf{0.51 \pm 0.12}$ & $\mathbf{0.56 \pm 0.22}$ & $\mathbf{2.35 \pm 1.34}$ \\
     \textbf{CSBA+UniKF} (N2) & $0.80 \pm 0.18$ & $1.32 \pm 0.53$ & $5.29 \pm 2.10$ \\
     \textbf{CSBA+UniKF} (N3) & $1.29 \pm 0.23$ & $2.55 \pm 1.02$ & $8.53 \pm 3.10$ \\
    \textbf{CSBA+UniKF} (N1\&N3) & $0.67 \pm 0.15$ & $0.76 \pm 0.30$ & $4.40 \pm 1.41$ \\
\midrule
     Single Detector (N1) & $0.72 \pm 0.17$ & $0.79 \pm 0.31$ & $3.14 \pm 1.09$ \\
     Single Detector (N2) & $1.10 \pm 0.21$ & $1.85 \pm 0.74$ & $6.95 \pm 1.71$ \\
     Single Detector (N3) & $1.73 \pm 0.29$ & $3.28 \pm 1.41$ & $11.08 \pm 2.45$ \\  
\bottomrule
\end{tabular}%
\label{tab:individualvsfusion}
\end{table}

~\cref{tab:individualvsfusion} presents a comparison of single-detector performance against our \textit{CSBA+UniKF} collaborative fusion approach under different noise conditions. The results demonstrate that collaboration consistently improves accuracy across all metrics.

Notably, \textit{CSBA+UniKF (Noise 1)} achieves the lowest mATE of $0.51\,m$ and mADE of $0.56\,m$, demonstrating superior localization and scale estimation compared to single detectors. Similarly, \textit{CSBA+UniKF (Noise 1 \& 3)} achieves a trade-off, with an mATE of $0.67\,m$, outperforming \textit{Single Detector (Noise 1)} ($0.72\,m$). However, its orientation accuracy is lower than \textit{Single Detector (Noise 1)}, with an mAOE of $4.40^\circ$ compared to $3.14^\circ$.

In contrast, single detectors perform worse in high-noise scenarios, as seen with \textit{Single Detector (Noise 3)}, which exhibits the highest errors. This highlights the sensitivity of individual detectors to perception noise, reinforcing the importance of uncertainty-aware collaboration, particularly in heterogeneous sensors' noise configurations.
\subsection{Comparison of SOTA and Proposed Metrics}
\label{subsec:comparison_metrics}

Tab. \ref{tab:after_before_metrics} compares state-of-the-art (SOTA) evaluation metrics with our proposed metrics, assessing fusion performance under the combination of noise levels: \textit{Noise 1} and \textit{Noise 3}. Notably, \textit{CSBA+UniKF} achieves identical results across both evaluation schemes, as it does not introduce false positives or false negatives, ensuring consistent performance. In contrast, while SOTA metrics indicate that \textit{DAIR-V2X-Late} achieves the lowest mADE of $0.61m$ and \textit{NMS-STD} the lowest mAOE of $3.12^\circ$, our proposed metrics reveal an increase in scale error for \textit{DAIR-V2X-Late} and a significant rise in orientation error for \textit{NMS-STD}. 

This discrepancy arises from a limitation in the SOTA evaluation pipeline, which only considers true positives in metrics computation, misrepresenting errors and penalizing methods that effectively reduce false positives. As a result, it favors approaches like \textit{NMS-STD}, which retain all detections without proper fusion. This highlights the need for a more comprehensive evaluation framework, such as the one we propose.

\begin{table}[h]
\caption{Comparison of fusion methods: \textbf{Ours (CSBA+UniKF)} vs. baselines NMS-STD~\cite{neubeck_efficient_2006}, DAIR-V2X-Late~\cite{yu2022dairv2x}. The metrics evaluate fusion performance under heterogeneous detectors with Noise 1 and Noise 3. mATE, mADE, and mAOE are respectively in meters, meters, and degrees.}
\centering
\scriptsize
\begin{tabular}{llccccc}
\toprule
Metrics & Method & mATE & mADE & mAOE & Precision & Recall \\
\midrule
\multirow{3}{*}{SOTA} 
    & \textbf{Ours} & $\mathbf{0.66}$ & $0.76$ & $4.39$ & $\mathbf{100}$ & $\mathbf{100}$ \\
    & \cite{neubeck_efficient_2006} & $0.72$ & $0.79$ & $\mathbf{3.12}$ & $58.55$ & $99.87$ \\
    & \cite{yu2022dairv2x} & $0.88$ & $\mathbf{0.61}$ & $6.40$ & $87.41$ & $\mathbf{100}$ \\
\midrule
\multirow{3}{*}{Proposed} 
    & \textbf{Ours} & $\mathbf{0.67}$ & $0.76$ & $\mathbf{4.38}$ & $\mathbf{100}$ & $\mathbf{100}$ \\
    & \cite{neubeck_efficient_2006} & $2.17$ & $1.32$ & $11.99$ & $58.55$ & $99.87$ \\
    & \cite{yu2022dairv2x} & $1.33$ & $\mathbf{0.74}$ & $8.13$ & $87.41$ & $\mathbf{100}$ \\ 
\bottomrule
\end{tabular}
\label{tab:after_before_metrics}
\end{table}

\subsection{Qualitative results}
\label{ssec:qualitative}
\begin{figure}[h!]
  \centering
  \setlength{\fboxrule}{0.8pt} 
  \vbox{
    \centering
    \fbox{
      \includegraphics[width=0.7\linewidth]{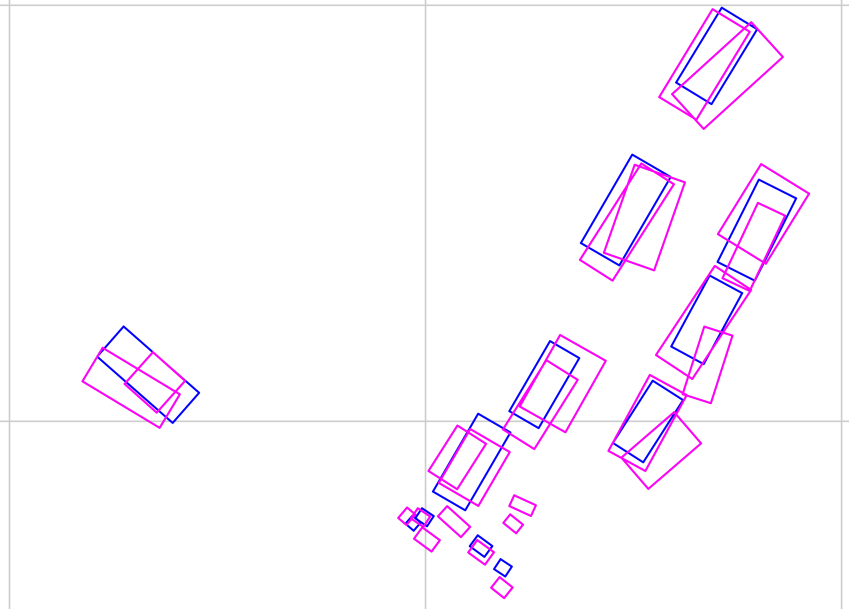}
    }
  }  
  \vspace{0.3cm} 
  \vbox{
    \centering
    \fbox{
      \includegraphics[width=0.7\linewidth]{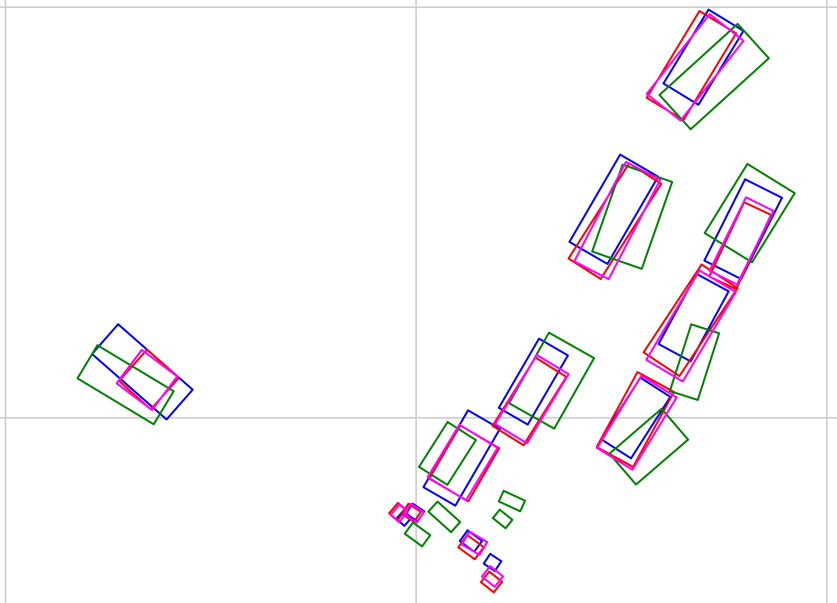}
    }
  } 
  \caption{Qualitative results of fusion algorithms of heterogeneous detectors with Noise 1 and Noise 3. The \textbf{top} figure shows results for \textit{NMS-STD}~\cite{neubeck_efficient_2006}, while the \textbf{bottom} figure illustrates the results of our proposed \textit{UniKF}. The visualizations include \textcolor{blue}{ground truth}, \textcolor{red}{Agent 1 detections}, \textcolor{green}{Agent 2 detections}, and \textcolor{magenta}{fused objects}. (Best viewed in color)}
  \label{fig:kfvsnms}
\end{figure}

Fig. ~\ref{fig:kfvsnms} illustrates the performance of \textit{NMS-STD} fusion method~\cite{neubeck_efficient_2006} compared to our proposed \textit{UniKF} under heterogeneous configuration of Noise 1 and Noise 3. We note that NMS-Based method does not actually perform a fusion and instead reatain all the reeceived detection. This is principally due to IOU-based asssociation method that require a universal threshold not adapted to all object categories and sizes which make it difficult to tune. In the other hand, UniKF perform the fusion for all the boxes and do not create data (false positive) or miss data (false negatives) as long as one of the sources provide a detection hence the $~100\%$ in Tab. \ref{tab:performance_metrics_all_dataset} recall and precision performance. The fusion is also correctly performed for small objects even if the translation and dimension noise are big. 

Fig. \ref{fig:kfvsnms} compares the performance of the baseline \textit{NMS-STD} fusion method~\cite{neubeck_efficient_2006} with our proposed \textit{UniKF} under a heterogeneous noise configuration (Noise 1 and Noise 3). \textit{NMS-STD} does not actively fuse detections but instead retains all received bounding boxes. This limitation arises from its IoU-based association mechanism, which relies on a fixed threshold—difficult to tune across varying object categories and scales. In contrast, \textit{UniKF} effectively performs fusion across all bounding boxes, avoiding both false positives and false negatives as long as at least one source provides a detection. This robustness leads to the nearly perfect $100\%$ recall and precision observed in Tab. \ref{tab:performance_metrics_all_dataset}. Moreover, \textit{UniKF} handles small objects reliably, even under significant translation and scale noise.

\subsection{Remaining Challenges}
A primary limitation of \textit{UniKF} lies in its underlying assumption that detection noise follows a Gaussian distribution. To assess robustness beyond the Gaussian assumption, we also tested our pipeline with other noise distributions (e.g., uniform, Laplacian and Student-t). While these introduce deviations from the Kalman filter's optimality conditions, our method still showed stable performance trends, suggesting good resilience to moderate distributional shifts. This behavior is expected, as stated in \cite{spall1995,maryak2004}: as long as the covariance used in the update step matches the true second moment of the noise, the Kalman filter remains the best linear unbiased estimator, regardless of the noise's exact distribution. For more extreme cases involving heavy-tailed, multimodal, or non-Gaussian noise, one could switch to particle filter.

Another key limitation of \textit{UniKF} is its inability to generate new detections. If an object is missed by all sensors, it cannot be recovered through fusion. Similarly, if two agents each produce a false positive that happens to be associated by the filter, the resulting fused output will also be a false positive.

\section{Conclusion and Perspectives}
\label{sec:Conclusion}
\raggedbottom
We introduced \textit{UniKF}, a novel uncertainty-aware late fusion framework, alongside a new evaluation protocol designed to \textbf{isolate and assess fusion performance independently}. Our proposed metrics offer a more accurate representation of fusion effectiveness, ensuring that efficient algorithms are not penalized.

Our results show that \textit{UniKF} consistently outperforms baseline methods across various noise levels and object categories, offering superior robustness to increasing uncertainty. Unlike heuristic-based fusion approaches that struggle with large uncertainties, \textit{UniKF} leverages uncertainty-aware fusion, ensuring stable and precise detection integration in multi-source collaborative perception.

While some challenges remain, the proposed approach offers a significant advancement in robust multi-source object fusion, providing a \textbf{bandwidth-efficient and detector-agnostic} framework that preserves intellectual property by eliminating the need for access to model features, unlike deep fusion methods. Future work will explore different motion models, although \cite{fadili2025} has demonstrated that the constant-velocity assumption is often the most effective. Additionally, we plan to incorporate explicit calibration offset modeling to mitigate potential sensor misalignment and expand our evaluation to include tracking-specific metrics and comparison with late tracking methods. Furthermore, we do not address classification confusion issues in this study, which can be particularly critical. As part of future work, we aim to integrate belief theory (e.g. Dempster-Shafer theory) to solve this issue. Finally, we aim to integrate detection into a complete perception pipeline, enabling direct comparisons with deep fusion approaches. 



\bibliographystyle{IEEEtran}
\bibliography{references}

\begin{thebibliography}{10}
\providecommand{\url}[1]{#1}
\csname url@samestyle\endcsname
\providecommand{\newblock}{\relax}
\providecommand{\bibinfo}[2]{#2}
\providecommand{\BIBentrySTDinterwordspacing}{\spaceskip=0pt\relax}
\providecommand{\BIBentryALTinterwordstretchfactor}{4}
\providecommand{\BIBentryALTinterwordspacing}{\spaceskip=\fontdimen2\font plus
\BIBentryALTinterwordstretchfactor\fontdimen3\font minus
  \fontdimen4\font\relax}
\providecommand{\BIBforeignlanguage}[2]{{%
\expandafter\ifx\csname l@#1\endcsname\relax
\typeout{** WARNING: IEEEtran.bst: No hyphenation pattern has been}%
\typeout{** loaded for the language `#1'. Using the pattern for}%
\typeout{** the default language instead.}%
\else
\language=\csname l@#1\endcsname
\fi
#2}}
\providecommand{\BIBdecl}{\relax}
\BIBdecl

\bibitem{han_collaborative_2023}
Y.~Han, H.~Zhang, H.~Li, Y.~Jin, C.~Lang, and Y.~Li, ``Collaborative perception
  in autonomous driving: Methods, datasets, and challenges,'' \emph{IEEE
  Intelligent Transportation Systems Magazine}, vol.~15, no.~6, 2023.

\bibitem{caillot_survey_2022}
A.~Caillot, S.~Ouerghi, P.~Vasseur, R.~Boutteau, and Y.~Dupuis, ``Survey on
  cooperative perception in an automotive context,'' \emph{IEEE Transactions on
  Intelligent Transportation Systems}, vol.~23, no.~9, pp. 14\,204--14\,223,
  2022.

\bibitem{huang_multi-modal_2022}
\BIBentryALTinterwordspacing
K.~Huang, B.~Shi, X.~Li, X.~Li, S.~Huang, and Y.~Li, ``Multi-modal sensor
  fusion for auto driving perception: A survey.'' [Online]. Available:
  \url{http://arxiv.org/abs/2202.02703}
\BIBentrySTDinterwordspacing

\bibitem{neubeck_efficient_2006}
\BIBentryALTinterwordspacing
A.~Neubeck and L.~Van~Gool, ``Efficient non-maximum suppression,'' in
  \emph{18th International Conference on Pattern Recognition
  ({ICPR}'06)}.\hskip 1em plus 0.5em minus 0.4em\relax {IEEE}, 2006, pp.
  850--855. [Online]. Available:
  \url{http://ieeexplore.ieee.org/document/1699659/}
\BIBentrySTDinterwordspacing

\bibitem{zhao_coopre_2024}
\BIBentryALTinterwordspacing
S.~Z. Zhao, H.~Xiang, C.~Xu, X.~Xia, B.~Zhou, and J.~Ma, ``{CooPre}:
  Cooperative pretraining for v2x cooperative perception.'' [Online].
  Available: \url{http://arxiv.org/abs/2408.11241}
\BIBentrySTDinterwordspacing

\bibitem{chen_co3_2022}
\BIBentryALTinterwordspacing
R.~Chen, Y.~Mu, R.~Xu, W.~Shao, C.~Jiang, H.~Xu, Z.~Li, and P.~Luo,
  ``{CO}{\textasciicircum}3: Cooperative unsupervised 3d representation
  learning for autonomous driving,'' publisher: {arXiv} Version Number: 2.
  [Online]. Available: \url{https://arxiv.org/abs/2206.04028}
\BIBentrySTDinterwordspacing

\bibitem{su_collaborative_2024}
\BIBentryALTinterwordspacing
S.~Su, S.~Han, Y.~Li, Z.~Zhang, C.~Feng, C.~Ding, and F.~Miao, ``Collaborative
  multi-object tracking with conformal uncertainty propagation,'' vol.~9,
  no.~4, pp. 3323--3330, conference Name: {IEEE} Robotics and Automation
  Letters. [Online]. Available:
  \url{https://ieeexplore.ieee.org/document/10430224}
\BIBentrySTDinterwordspacing

\bibitem{liu_when2com_2020}
\BIBentryALTinterwordspacing
Y.-C. Liu, J.~Tian, N.~Glaser, and Z.~Kira, ``When2com: Multi-agent perception
  via communication graph grouping.'' [Online]. Available:
  \url{http://arxiv.org/abs/2006.00176}
\BIBentrySTDinterwordspacing

\bibitem{li_v2x-dgw_2024}
\BIBentryALTinterwordspacing
B.~Li, J.~Li, X.~Liu, R.~Xu, Z.~Tu, J.~Guo, X.~Li, and H.~Yu, ``V2x-{DGW}:
  Domain generalization for multi-agent perception under adverse weather
  conditions.'' [Online]. Available: \url{http://arxiv.org/abs/2403.11371}
\BIBentrySTDinterwordspacing

\bibitem{qiao_cobevfusion_2023}
\BIBentryALTinterwordspacing
D.~Qiao and F.~Zulkernine, ``{CoBEVFusion}: Cooperative perception with
  {LiDAR}-camera bird's-eye view fusion.'' [Online]. Available:
  \url{http://arxiv.org/abs/2310.06008}
\BIBentrySTDinterwordspacing

\bibitem{zimmer_tumtraf_2024}
\BIBentryALTinterwordspacing
W.~Zimmer, G.~A. Wardana, S.~Sritharan, X.~Zhou, R.~Song, and A.~C. Knoll,
  ``{TUMTraf} v2x cooperative perception dataset.'' [Online]. Available:
  \url{http://arxiv.org/abs/2403.01316}
\BIBentrySTDinterwordspacing

\bibitem{abdali2025}
B.~Abdali, Q.~Picard, and M.~Fadili, ``Data optimization strategies for
  collaborative perception,'' \emph{Electronic Imaging}, vol.~37, pp. 1--5,
  2025.

\bibitem{bodla_soft-nms_2017}
\BIBentryALTinterwordspacing
N.~Bodla, B.~Singh, R.~Chellappa, and L.~S. Davis, ``Soft-{NMS} — improving
  object detection with one line of code,'' in \emph{2017 {IEEE} International
  Conference on Computer Vision ({ICCV})}.\hskip 1em plus 0.5em minus
  0.4em\relax {IEEE}, pp. 5562--5570. [Online]. Available:
  \url{http://ieeexplore.ieee.org/document/8237855/}
\BIBentrySTDinterwordspacing

\bibitem{xu_model-agnostic_2023}
R.~Xu, W.~Chen, H.~Xiang, X.~Xia, L.~Liu, and J.~Ma, ``Model-agnostic
  multi-agent perception framework,'' in \emph{2023 IEEE International
  Conference on Robotics and Automation (ICRA)}, 2023, pp. 1471--1478.

\bibitem{shen_competitive_2021}
\BIBentryALTinterwordspacing
F.~Shen, X.~He, M.~Wei, and Y.~Xie, ``A competitive method to {VIPriors} object
  detection challenge,'' version Number: 1. [Online]. Available:
  \url{https://arxiv.org/abs/2104.09059}
\BIBentrySTDinterwordspacing

\bibitem{zimmer_infradet3d_2023}
\BIBentryALTinterwordspacing
W.~Zimmer, J.~Birkner, M.~Brucker, H.~T. Nguyen, S.~Petrovski, B.~Wang, and
  A.~C. Knoll, ``{InfraDet}3d: Multi-modal 3d object detection based on
  roadside infrastructure camera and {LiDAR} sensors.'' [Online]. Available:
  \url{http://arxiv.org/abs/2305.00314}
\BIBentrySTDinterwordspacing

\bibitem{yu_dair-v2x_2022}
\BIBentryALTinterwordspacing
H.~Yu, Y.~Luo, M.~Shu, Y.~Huo, Z.~Yang, Y.~Shi, Z.~Guo, H.~Li, X.~Hu, J.~Yuan,
  and Z.~Nie, ``{DAIR}-v2x: A large-scale dataset for vehicle-infrastructure
  cooperative 3d object detection.'' [Online]. Available:
  \url{http://arxiv.org/abs/2204.05575}
\BIBentrySTDinterwordspacing

\bibitem{picard_decentralized}
\BIBentryALTinterwordspacing
Q.~Picard, M.~Morice, M.~Fadili, and S.~Pechberti, ``{Decentralized perception
  system with multiple viewpoints},'' Oct. 2024, working paper or preprint.
  [Online]. Available: \url{https://hal.science/hal-04744167}
\BIBentrySTDinterwordspacing

\bibitem{wang_survey_2023}
\BIBentryALTinterwordspacing
L.~Wang, X.~Zhang, Z.~Song, J.~Bi, G.~Zhang, H.~Wei, L.~Tang, L.~Yang, J.~Li,
  C.~Jia, and L.~Zhao, ``\BIBforeignlanguage{en}{Multi-{Modal} {3D} {Object}
  {Detection} in {Autonomous} {Driving}: {A} {Survey} and {Taxonomy}},''
  \emph{\BIBforeignlanguage{en}{IEEE Transactions on Intelligent Vehicles}},
  vol.~8, no.~7, pp. 3781--3798, Jul. 2023. [Online]. Available:
  \url{https://ieeexplore.ieee.org/document/10093116/}
\BIBentrySTDinterwordspacing

\bibitem{wang_multi-sensor_2024}
\BIBentryALTinterwordspacing
X.~Wang, K.~Li, and A.~Chehri, ``Multi-{Sensor} {Fusion} {Technology} for {3D}
  {Object} {Detection} in {Autonomous} {Driving}: {A} {Review},'' \emph{IEEE
  Transactions on Intelligent Transportation Systems}, vol.~25, no.~2, pp.
  1148--1165, Feb. 2024, conference Name: IEEE Transactions on Intelligent
  Transportation Systems. [Online]. Available:
  \url{https://ieeexplore.ieee.org/abstract/document/10265760}
\BIBentrySTDinterwordspacing

\bibitem{MV3D2017}
X.~Chen, H.~Ma, J.~Wan, B.~Li, and T.~Xia, ``Multi-view 3d object detection
  network for autonomous driving,'' in \emph{Proceedings of the IEEE Conference
  on Computer Vision and Pattern Recognition (CVPR)}, July 2017.

\bibitem{avod_2018}
J.~Ku, M.~Mozifian, J.~Lee, A.~Harakeh, and S.~L. Waslander, ``Joint 3d
  proposal generation and object detection from view aggregation,'' in
  \emph{2018 IEEE/RSJ International Conference on Intelligent Robots and
  Systems (IROS)}, 2018, pp. 1--8.

\bibitem{pang2020clocs}
S.~Pang, D.~Morris, and H.~Radha, ``Clocs: Camera-lidar object candidates
  fusion for 3d object detection,'' 2020.

\bibitem{pang2022}
------, ``Fast-clocs: Fast camera-lidar object candidates fusion for 3d object
  detection,'' in \emph{Proceedings of the IEEE/CVF Winter Conference on
  Applications of Computer Vision (WACV)}, January 2022, pp. 187--196.

\bibitem{li_delving_2023}
\BIBentryALTinterwordspacing
H.~Li, C.~Sima, J.~Dai, W.~Wang, L.~Lu, H.~Wang, J.~Zeng, Z.~Li, J.~Yang,
  H.~Deng, H.~Tian, E.~Xie, J.~Xie, L.~Chen, T.~Li, Y.~Li, Y.~Gao, X.~Jia,
  S.~Liu, J.~Shi, D.~Lin, and Y.~Qiao, ``Delving into the devils of
  bird's-eye-view perception: A review, evaluation and recipe,'' pp. 1--20,
  2023, conference Name: {IEEE} Transactions on Pattern Analysis and Machine
  Intelligence. [Online]. Available:
  \url{https://ieeexplore.ieee.org/abstract/document/10321736}
\BIBentrySTDinterwordspacing

\bibitem{castanedo_review_2013}
F.~Castanedo, ``A review of data fusion techniques,'' vol. 2013, p. 704504.

\bibitem{su_uncertainty_2023}
S.~Su, Y.~Li, S.~He, S.~Han, C.~Feng, C.~Ding, and F.~Miao, ``Uncertainty
  quantification of collaborative detection for self-driving,'' in \emph{2023
  IEEE International Conference on Robotics and Automation (ICRA)}.\hskip 1em
  plus 0.5em minus 0.4em\relax IEEE, 2023, pp. 5588--5594.

\bibitem{mun_uncertainty_2023}
\BIBentryALTinterwordspacing
J.~Mun and H.~Choi, ``Uncertainty prediction for monocular 3d object
  detection,'' vol.~23, no.~12, p. 5395, number: 12 Publisher:
  Multidisciplinary Digital Publishing Institute. [Online]. Available:
  \url{https://www.mdpi.com/1424-8220/23/12/5395}
\BIBentrySTDinterwordspacing

\bibitem{DMSTrack}
H.-K. Chiu, C.-Y. Wang, M.-H. Chen, and S.~F. Smith, ``Probabilistic 3d
  multi-object cooperative tracking for autonomous driving via differentiable
  multi-sensor kalman filter,'' in \emph{2024 IEEE International Conference on
  Robotics and Automation (ICRA)}, 2024.

\bibitem{fadili2025}
\BIBentryALTinterwordspacing
M.~Fadili, L.~Lecrosnier, S.~Pechberti, and R.~Khemmar, ``Weighted
  least-squares multi-detection fusion and kalman filter-based tracking for
  collaborative perception systems.'' [Online]. Available:
  \url{https://hal.science/hal-04910986}
\BIBentrySTDinterwordspacing

\bibitem{harimohan2019}
H.~Jha, V.~Lodhi, and D.~Chakravarty, ``Object detection and identification
  using vision and radar data fusion system for ground-based navigation,'' in
  \emph{2019 6th International Conference on Signal Processing and Integrated
  Networks (SPIN)}, 2019, pp. 590--593.

\bibitem{geiger_vision_2013}
\BIBentryALTinterwordspacing
A.~Geiger, P.~Lenz, C.~Stiller, and R.~Urtasun, ``Vision meets robotics: The
  {KITTI} dataset,'' vol.~32, no.~11, pp. 1231--1237. [Online]. Available:
  \url{http://journals.sagepub.com/doi/10.1177/0278364913491297}
\BIBentrySTDinterwordspacing

\bibitem{caesar_nuscenes_2020}
\BIBentryALTinterwordspacing
H.~Caesar, V.~Bankiti, A.~H. Lang, S.~Vora, V.~E. Liong, Q.~Xu, A.~Krishnan,
  Y.~Pan, G.~Baldan, and O.~Beijbom, ``{nuScenes}: A multimodal dataset for
  autonomous driving,'' 2020. [Online]. Available:
  \url{http://arxiv.org/abs/1903.11027}
\BIBentrySTDinterwordspacing

\bibitem{yu2022dairv2x}
C.~Yu, Y.~Li, and R.~Zhang, ``Dair-v2x: A large-scale real-world v2i
  collaborative perception dataset,'' in \emph{Proceedings of the IEEE/CVF
  Conference on Computer Vision and Pattern Recognition (CVPR)}, 2022.

\bibitem{LeUncertainty2018}
M.~T. Le, F.~Diehl, T.~Brunner, and A.~Knoll, ``Uncertainty estimation for deep
  neural object detectors in safety-critical applications,'' in \emph{2018 21st
  International Conference on Intelligent Transportation Systems (ITSC)}, 2018,
  pp. 3873--3878.

\bibitem{burkardt_truncated_2009}
J.~Burkardt, ``The truncated normal distribution.''

\bibitem{baisa_derivation_2020}
\BIBentryALTinterwordspacing
N.~L. Baisa, ``Derivation of a constant velocity motion model for visual
  tracking.'' [Online]. Available: \url{http://arxiv.org/abs/2005.00844}
\BIBentrySTDinterwordspacing

\bibitem{xu2022opv2v}
R.~Xu, T.~Wang, and Y.~Chen, ``Opv2v: Opencda-based v2v collaborative
  perception dataset,'' in \emph{Proceedings of the IEEE International
  Conference on Robotics and Automation (ICRA)}, 2022.

\bibitem{hu2022v2xset}
Z.~Hu, F.~Liu, and Y.~Shen, ``V2xset: A large-scale open simulation dataset for
  v2x perception,'' in \emph{Proceedings of the European Conference on Computer
  Vision (ECCV)}, 2022.

\bibitem{li2021v2xsim}
Q.~Li, T.~Wang, and Y.~Chen, ``V2x-sim: Multi-agent collaborative perception
  dataset,'' \emph{IEEE Robotics and Automation Letters}, 2021.

\bibitem{gawlikowski_2023}
\BIBentryALTinterwordspacing
J.~Gawlikowski, C.~R.~N. Tassi, M.~Ali, J.~Lee, M.~Humt, J.~Feng, A.~Kruspe,
  R.~Triebel, P.~Jung, R.~Roscher, M.~Shahzad, W.~Yang, R.~Bamler, and X.~X.
  Zhu, ``\BIBforeignlanguage{en}{A survey of uncertainty in deep neural
  networks},'' \emph{\BIBforeignlanguage{en}{Artificial Intelligence Review}},
  vol.~56, no.~S1, pp. 1513--1589, Oct. 2023. [Online]. Available:
  \url{https://link.springer.com/10.1007/s10462-023-10562-9}
\BIBentrySTDinterwordspacing

\bibitem{weng_ab3dmot_2020}
\BIBentryALTinterwordspacing
X.~Weng, J.~Wang, D.~Held, and K.~Kitani, ``{AB}3dmot: A baseline for 3d
  multi-object tracking and new evaluation metrics.'' [Online]. Available:
  \url{http://arxiv.org/abs/2008.08063}
\BIBentrySTDinterwordspacing

\bibitem{yang2023v2v4real}
Y.~Yang, X.~Ma, and R.~Wang, ``V2v4real: A real-world multimodal collaborative
  perception dataset for vehicle-to-vehicle interaction,'' in \emph{Proceedings
  of the IEEE/CVF Conference on Computer Vision and Pattern Recognition
  (CVPR)}, 2023.

\bibitem{giou2019}
\BIBentryALTinterwordspacing
H.~Rezatofighi, N.~Tsoi, J.~Gwak, A.~Sadeghian, I.~Reid, and S.~Savarese,
  ``Generalized intersection over union: A metric and a loss for bounding box
  regression.'' [Online]. Available: \url{http://arxiv.org/abs/1902.09630}
\BIBentrySTDinterwordspacing

\bibitem{spall1995}
\BIBentryALTinterwordspacing
J.~C. Spall, ``The kantorovich inequality for error analysis of the kalman
  filter with unknown noise distributions,'' \emph{Automatica}, vol.~31,
  no.~10, pp. 1513--1517, 1995. [Online]. Available:
  \url{https://www.sciencedirect.com/science/article/pii/0005109895000699}
\BIBentrySTDinterwordspacing

\bibitem{maryak2004}
J.~Maryak, J.~Spall, and B.~Heydon, ``Use of the kalman filter for inference in
  state-space models with unknown noise distributions,'' \emph{IEEE
  Transactions on Automatic Control}, vol.~49, no.~1, pp. 87--90, 2004.

\end{thebibliography}

\relax 

\end{document}